\documentclass[10pt,twocolumn,letterpaper]{article}

\usepackage{cvpr}              % To produce the CAMERA-READY version
\usepackage[dvipsnames]{xcolor}

\newcommand{\TODO}[1]{\textbf{\color{red}[TODO: #1]}}
\renewcommand{\TODO}[1]{}

\definecolor{cvprblue}{rgb}{0.21,0.49,0.74}
\usepackage[pagebackref,breaklinks,colorlinks,citecolor=cvprblue]{hyperref}
\usepackage{multirow}
\usepackage[misc]{ifsym}
\usepackage[accsupp]{axessibility}
\def\confName{CVPR}
\def\confYear{2024}

\title{Low-Rank Knowledge Decomposition for Medical Foundation Models}

\author{Yuhang Zhou\textsuperscript{1,2}, 
 Haolin Li\textsuperscript{2,3}, 
 Siyuan Du\textsuperscript{2,3}, 
 Jiangchao Yao\textsuperscript{1,2,\Letter}, 
 Ya Zhang\textsuperscript{1,2}, 
 Yanfeng Wang\textsuperscript{1,2,\Letter}\\
\textsuperscript{1}Cooperative Medianet Innovation Center, Shanghai Jiao Tong University \\
\textsuperscript{2}Shanghai Artificial Intelligence Laboratory \quad \textsuperscript{3}Fudan University \\
\tt\small{\{zhouyuhang, Sunarker, ya\_zhang, wangyanfeng622\}@sjtu.edu.cn}, \\ \tt\small{\{dusiyuan, lihaolin\}@pjlab.org.cn}}

\begin{document}
\maketitle
\begin{abstract}
The popularity of large-scale pre-training has promoted the development of medical foundation models. However, some studies have shown that although foundation models exhibit strong general feature extraction capabilities, their performance on specific tasks is still inferior to task-specific methods. 
In this paper, we explore a new perspective called ``Knowledge Decomposition'' to improve the performance on specific medical tasks, which deconstruct the foundation model into multiple lightweight expert models, each dedicated to a particular task, with the goal of improving specialization while concurrently mitigating resource expenditure.
To accomplish the above objective, we design a novel framework named Low-Rank Knowledge Decomposition (LoRKD), which explicitly separates graidents by incorporating low-rank expert modules and the efficient knowledge separation convolution. 
Extensive experimental results demonstrate that the decomposed models perform well in terms of performance and transferability, even surpassing the original foundation models. 
Source code is available at: \url{https://github.com/MediaBrain-SJTU/LoRKD}

\end{abstract}    
\section{Introduction}
\label{sec:intro}
Foundation models pre-trained on large-scale and diverse datasets, have been proven to possess powerful general feature extraction capabilities and can handle various tasks~\cite{bommasani2021opportunities,  sellergren2022simplified}. 
However, some studies~\cite{glocker2023risk, huang2023segment,wu2023can} have indicated that the performance of foundation models is still inferior to task-specific methods, suggesting that current foundation models are unable to simultaneously guarantee both generality and specialization. Moreover, with the gradual expansion of data scale and model capacity, the deployment costs of future foundation models may become exorbitant.  To address them, we proposes a new perspective called knowledge decomposition, aiming to offer potential solutions for the practical application of medical foundation models.

\begin{figure}[t]
    \centering
    \includegraphics[width=\linewidth]{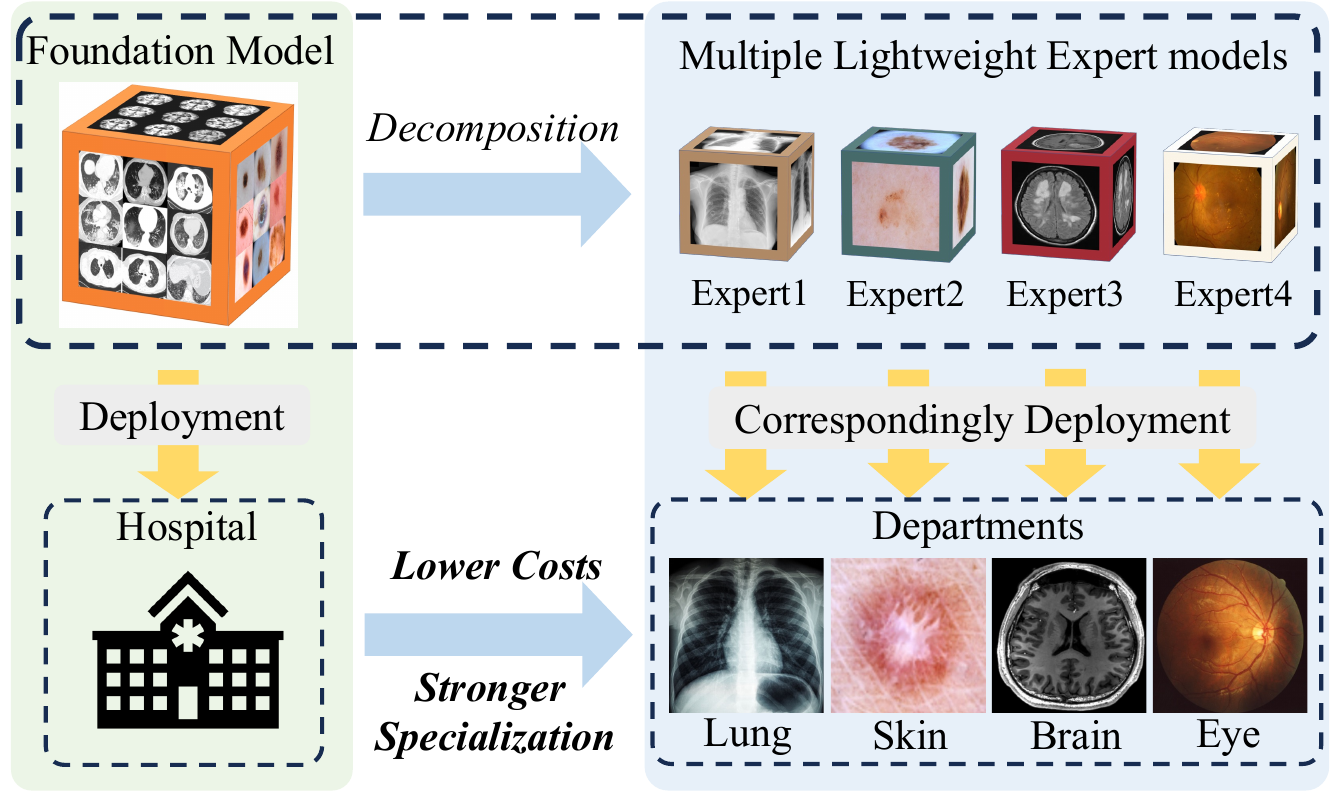}
    \caption{Knowledge decomposition is employed to break down the foundation model into multiple lightweight expert models, each dedicated to a specific domain. The goal of this paradigm is to improve the specialization of deployment models within a specific domain, while simultaneously reducing deployment costs.}
    \label{fig:background}
    \vspace{-0.2cm}
\end{figure}

The purpose of knowledge decomposition is to break down the foundation model into multiple lightweight expert models, where each expert model focuses solely on a specific domain, such as a department within a hospital (as shown in Figure~\ref{fig:background}). 
The resulting expert models, compared to the original foundation model, gain stronger specialization and lower deployment costs. 
In order to preserve both the performance and transferability of decomposed models, we need to inject task-specific knowledge and common knowledge into each expert model correspondingly, which is highly challenging.
To the best of our knowledge, there has been no research conducted in the medical field on how to decompose a foundation model into multiple expert models. However, recently in the field of natural images, KF~\cite{yang2022factorizing} has made preliminary explorations into this issue. KF decomposes the pre-trained model into a common knowledge network (CKN) and multiple task-specific networks (TSNs) by manipulating the mutual information between models. After decomposition, the CKN can be combined with each TSN to form task-specific expert models. 
However, in our experiments, we find that the effectiveness of this method in medical scenarios is not significant.

In this paper, we propose a novel method for knowledge decomposition of medical foundation models called Low-Rank Knowledge Decomposition (LoRKD). 
LoRKD consists of two main components: low-rank expert modules and the efficient knowledge separation convolution. The former provides multiple parameter-efficient task-specific knowledge carriers for each convolution, which effectively controls the introduction of parameters while ensuring sufficient feature representation capability. 
The latter provides an efficient implementation method for expert knowledge separation at the convolutional level, allowing gradients to be separated into the corresponding expert modules in a single forward propagation, while accumulating them in the shared backbone.
This ensures that each expert module learns task-specific knowledge while the shared backbone learns common knowledge. 
After decomposition, we can integrate the task-specific expert modules and the shared backbone through parameter fusion, ensuring model performance and transferability without increasing additional parameters. 
Furthermore, benefiting from the training pattern and parameter fusion mechanism, our decomposed model can easily switch task knowledge across different domains.
The performance comparison on three pre-training datasets and seven downstream datasets demonstrates the effectiveness of LoRKD. A large number of analytical experiments further showcase the advantages of LoRKD from different perspectives.
In a nutshell, our contributions can be summarized as the following:
\begin{itemize}
\item We introduce knowledge decomposition to broaden application of medical foundation models, which decomposes models into multiple lightweight experts to reduce costs and enhance specialization. The incorporation of this novel perspective offers potential solutions for the practical implementation of medical foundation models.
\item We design a new method LoRKD, which consists of two components: low-rank expert modules and the efficient knowledge separation convolution. LoRKD injects task-specific knowledge into the corresponding expert modules through efficient explicit gradient separation.
\item A significant number of experiments and analyses have demonstrated the effectiveness of LoRKD and the potential of knowledge decomposition. 
\end{itemize}

\section{Related work}
\label{sec:related_work}

\par{\noindent \bf Knowledge Distillation.}
Knowledge distillation (KD)~\cite{hinton2015distilling} is an effective knowledge transfer method, which can be  categorized into two groups: logits-based distillation~\cite{furlanello2018born,  mirzadeh2020improved, yang2023knowledge} and feature-based distillation~\cite{romero2014fitnets, tung2019similarity, ahn2019variational, park2019relational, zagoruyko2016paying, touvron2021training, tian2019contrastive, fang2021seed}.
The former encourages students to mimic the softmax outputs of teacher models, while the latter encourages students to mimic the intermediate-level features from the hidden layers of teacher models.
Diﬀerent from these methods that focus on transferring complete knowledge, 
our goal is to decompose knowledge into different expert models.

\par{\noindent \bf Multi-Task Learning.}
Multi-task learning (MTL)~\cite{caruana1993multitask, crawshaw2020multi, ruder2017overview} aims to train a unified model to solve multiple distinct but related tasks ~\cite{caruana1997multitask, zhang2021survey, senushkin2023independent, fernando2022mitigating, sener2018multi, liu2019end}. Therefore, the main focus of MTL is to train better general feature extractor. For example, MoCo~\cite{fernando2022mitigating} addresses the convergence issue of Multi-Gradient Descent Algorithm (MGDA)~\cite{desideri2012multiple} to ensure convergence to Pareto optima. Aligned-MTL~\cite{senushkin2023independent} stabilizes the training process by aligning the principal components of the gradient matrix. 
However, tasks within the context of foundation models often display significant diversity, and solely pursuing common knowledge may not be appropriate.
In summary, MTL focuses on extracting shared knowledge from relevant tasks, while knowledge decomposition emphasizes separating task-specific knowledge from the foundation model trained on diverse tasks.

\par{\noindent \bf Knowledge Decomposition.} 
Different from the previous disentangled representation learning that are usually done through adversarial learning~\cite{tran2017disentangled, chen2016infogan, liu2018multi, mathieu2016disentangling} or variational auto-encoder~\cite{burgess2018understanding, higgins2016beta, kim2018disentangling}, the goal of knowledge decomposition is to break down the pre-trained foundation model into multiple task-specific experts.
Recently, in the field of natural images, KF~\cite{yang2022factorizing} has explored knowledge decomposition by promoting modularization of knowledge through optimizing mutual information loss~\cite{hjelm2018learning, lowe2019greedy,oord2018representation}. 
It decomposes a pre-trained model into a common knowledge network and multiple task-specific networks. 
In this paper, we conduct the first exploration of knowledge decomposition in the medical field and propose a novel approach that not only better controls the number of parameters  but also attains a more advanced level of performance and transferability.

\section{Proposed Method}
\begin{figure*}[t]
    \centering
    \includegraphics[width=0.96\linewidth]{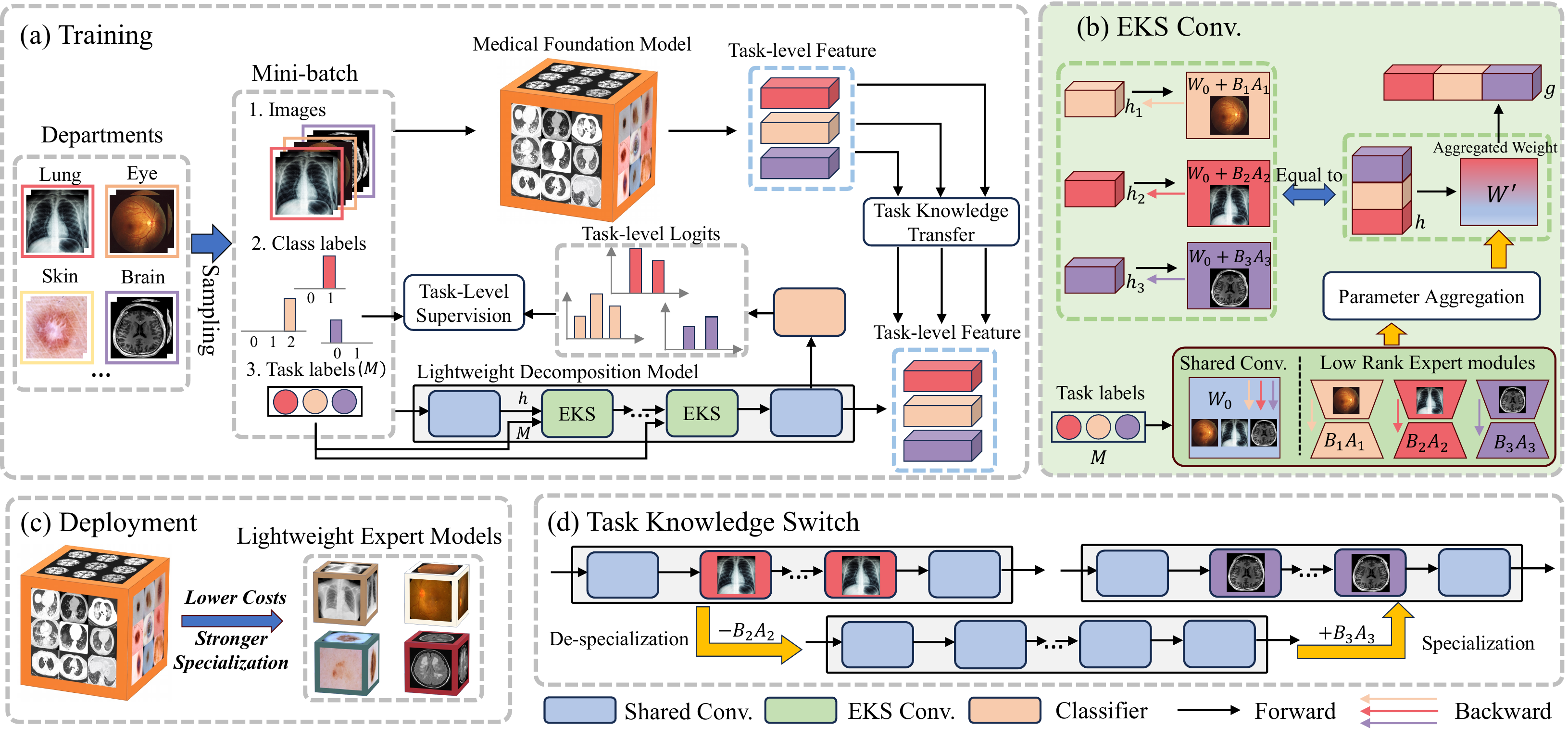}
    \caption{The overview of LoRKD. We introduce low-rank expert modules to control the number of parameters and efficient knowledge separation convolution to achieve computationally efficient explicit gradient separation. The decomposed  models can replace the original foundation model in specific domains, and can switch task knowledge conveniently between different departments.
    }
    \vspace{-0.3cm}
    \label{fig:method}
\end{figure*}

Given a medical foundation model $F_p$ pre-trained on a broad range of data, our goal is to decompose $F_p$ into $T$ lightweight expert models ${F_1,..., F_T}$ that can be deployed to $T$ different medical departments instead of using $F_p$. Our lightweight decomposition model comprises a shared backbone $F_s$ and $T$ expert modules $\{E_1, ..., E_T\}$ during training. 
To achieve efficient knowledge decomposition, we propose low-rank expert modules and efficient knowledge separation convolution which will be described in detail below. An overview of our method can be seen in Figure~\ref{fig:method}.

\subsection{Low Rank Expert Modules}
Considering the limited computational resources and the scalability of the number of tasks, expert modules, as carriers of task-specific knowledge, need to strike a balance between the number of parameters and the feature representation capability. 
LoRA~\cite{hu2021lora}, as a commonly used fine-tune method in foundation models, has been proven to be parameter-efficient~\cite{zhang2023lora,valipour2022dylora}.
Inspired by this, we propose to use a similar low-rank structure as the carriers for knowledge decomposition, named low rank expert modules.

Given a shared convolution $\mathbf{W_0}\in \mathbb{R}^{C^{\text{out}}\times C^{\text{in}}\times k\times k}$ in $F_s$, where $C^{\text{out}}, C^{\text{in}}, k$ represent the number of output channels, the number of input channels, and the kernel size respectively, we configure two low-rank factors $\mathbf{B_t}\in \mathbb{R}^{C^{\text{out}}k\times rk}$ and $\mathbf{A_t} \in \mathbb{R}^{rk\times C^{\text{in}}k}$ for $t$-th expert, where $r$ represents the rank. 
As a result, for the features belonging to the $t$-th task, original convolution operation $g_t = \mathbf{W_0}h_t$ can be transformed into $g_t = (\mathbf{W_0}+\mathbf{B_t}\mathbf{A_t}) h_t$, where, for brevity, we omit the reshape operation, and $h_t$, $g_t$ represent the input features and output features respectively.

It is worth noting that, unlike previous scenarios  where $\mathbf{W_0}$ remains fixed in LoRA, in our knowledge decomposition scenario, $\mathbf{W_0}$, as a carrier of common knowledge, requires updating along with the low-rank factors.

\subsection{Efficient Knowledge Separation Convolution}
To achieve knowledge decomposition, we propose explicit gradient separation as our solution. This approach requires each expert module to compute gradients solely for its corresponding task, enabling the acquisition of task-specific knowledge.  Simultaneously, the shared backbone collects gradients from all tasks to facilitate the acquisition of common knowledge among tasks. 
However, when a mini-batch of data contains $T$ tasks, the convolution operation becomes $T$ times $g_t = (\mathbf{W_0}+\mathbf{B_t}\mathbf{A_t}) h_t$, where $t\in \{1,...,T\}$. The $T$ times forward propagation significantly increases the training time, especially when a large number of tasks needed to be decomposed. To address this, we propose Efficient Knowledge Separation Convolution (EKS Conv.).

In order to clarify our improvements in convolution, we first review the standard convolution operation. In a deep convolutional neural network, the input features of each convolution can be represented as $h \in \mathbb{R}^{B \times C^{\text{in}} \times H \times W}$, where $B, H, W$ represent the sample number of a mini-batch, the height and width of the feature maps, respectively. 
If the kernel size of the convolution is $k$ and the stride is 1, each output feature unit $o_{ij} \in \mathbb{R}^{B \times C^{\text{out}}}$ in output features $g \in \mathbb{R}^{B \times C^{\text{out}} \times H \times W}$ can be represented as
\begin{align}
o_{ij}=\sum_{m=0}^{k-1}\sum_{n=0}^{k-1} h_{(i+m)(j+n)} \cdot \omega_{mn}, \nonumber
\end{align}
where $i \in \{ 1, ..., H \}$ , $j \in \{ 1, ..., W \}$ , and $h_{(i+m)(j+n)} \in \mathbb{R}^{B\times C^{\text{in}}}$ represents the units of the input feature map $h$, and $\omega_{mn} \in \mathbb{R}^{C^{\text{in}} \times C^{\text{out}}}$ represents the convolution weights.

Our EKS Convolution improves upon the traditional convolution operation by enabling gradient separation to be achieved in a single forward propagation, regardless of the number of tasks. EKS Convolution avoids the computational overhead of duplicating data input for each convolution, greatly enhancing training efficiency.
Specifically, for each EKS Convolution, in addition to the input feature map $h$, the task label $\mathbf{M} \in \mathbb{R}^{B\times  T}$, corresponding to the mini-batch is also simultaneously inputted as a reference for subsequent parameter aggregation, and $\mathbf{M}$ is a one-hot vector. 
Then, the output features can be computed by
\begin{equation}
\setlength{\abovedisplayskip}{3pt}
\setlength{\belowdisplayskip}{3pt}
    \begin{aligned}
    g&=g_1\cup\cdots\cup g_t \cup\cdots\cup g_T\\
    g_t&=(\mathbf{W_0}+\mathbf{B_t}\mathbf{A_t})h_t =(\mathbf{W_0}+\mathbf{B_t}\mathbf{A_t})\mathbf{M_t}h, 
\end{aligned}
\label{eq:forward}
\end{equation}
where $\cup$ means the concatenation operation, $h_t $ represents the set of $B^t$ features in $h$ that correspond to the $t$-th task, and  $\mathbf{M_t}$ is an index matrix that indicates which features in $h$ belong to the $t$-th task. 
To avoid redundant convolutional operations, we propose parameter aggregation, where the parameters used in the current iteration are aggregated into $\mathbf{W^{\prime}}$ according to $\mathbf{M}$. This ensures that the number of forward propagation is always equal to 1, and the operation $g=\mathbf{W^{\prime}}h$ is equivalent to the Eqn.~\eqref{eq:forward}. Specifically, the operation of the Eqn.~\eqref{eq:forward} can be converted by 
\begin{equation}
\setlength{\abovedisplayskip}{3pt}
\setlength{\belowdisplayskip}{3pt}
\begin{aligned}
    g & =(\mathbf{W_0}+\mathbf{B_1}\mathbf{A_1})h_1\cup\cdots\cup (\mathbf{W_0}+\mathbf{B_T}\mathbf{A_T})h_T \\
    &=(\mathbf{W_0}+\sum\nolimits_{i=1}^{T} (\mathbf{\widetilde{BA}}\odot \mathbf{M})_i)h =\mathbf{W^{\prime}} h,
\end{aligned}
\nonumber
\end{equation}
where $\mathbf{\widetilde{BA}} \in \mathbb{R}^{1\times T\times C^{\text{out}}\times C^{\text{in}} \times k \times k}$ contains the weight of all low-rank expert modules, which can be obtained by
\begin{equation}
    \mathbf{\widetilde{BA}} = \mathbf{B_1}\mathbf{A_1} \cup ...\cup \mathbf{B_t}\mathbf{A_t}\cup...\cup \mathbf{B_T}\mathbf{A_T}.\nonumber
\end{equation}
$\odot$ represents the Hadamard product, and $\mathbf{\widetilde{BA}}\odot \mathbf{M} \in \mathbb{R}^{B\times T\times C^{\text{out}}\times C^{\text{in}} \times k \times k}$ represents the configuration of low-rank expert modules for the each input feature and $i$ corresponds to the second dimension of $(\mathbf{\widetilde{BA}}\odot \mathbf{M})$. The weight of shared convolution $\mathbf{W_0}$ is applied to all tasks.
In this way, we obtain the aggregated weight $\mathbf{W^{\prime}} \in \mathbb{R}^{B\times C^{\text{out}}\times C^{\text{in}} \times k \times k}$, which is  equivalent to Eqn.~\eqref{eq:forward} but requires only a single forward propagation.

Another challenge associated with it is that $\mathbf{W^{\prime}}$ has 5 dimensions, unlike traditional convolutions which typically have 4 dimensions. To ensure compatibility with existing deep learning libraries, we have borrowed the concept of group convolution (GConv)~\cite{krizhevsky2012imagenet}.
Specifically, we set the group number to $B$ and $\gamma\in \{1,...,B\}$. Then, we reshape $h$ to $h \in \mathbb{R}^{1\times BC^{\text{in}}\times H\times W}$ and reshape $\mathbf{W^{\prime}}$ to $\mathbf{W^{\prime}}\in \mathbb{R}^{BC^{\text{out}}\times C^{\text{in}}\times k\times k}$. Consequently, each output feature unit $o_{ij}$ in $g$ can be computed by 
\begin{equation}
\setlength{\abovedisplayskip}{3pt}
\setlength{\belowdisplayskip}{3pt}
\begin{aligned}
    o_{ij}=o_{ij}^1\cup\cdots\cup o_{ij}^\gamma\cup\cdots\cup o_{ij}^B \\
o_{ij}^\gamma=\sum_{m=0}^{k-1}\sum_{n=0}^{k-1}h_{(i+m)(j+n)}^\gamma  \cdot \omega_{mn}^\gamma, 
\end{aligned}
\label{gconv}
\end{equation}
where $h_{(i+m)(j+n)}^\gamma$ and $\omega_{mn}^\gamma$ represent the reshaped version.
Eqn.~\eqref{gconv} is a standard form of group convolution, which can be easily implemented in existing deep learning libraries.
{A notation table can be found in the supplementary materials.}

\subsection{Loss Function}
In order to transfer the knowledge from the medical foundation model into the lightweight decompostion model, we introduce a task knowledge transfer loss denoted as $\mathcal{L}_{\mathrm{transfer}}$.
Specifically, for a mini-batch of training data $\{(x_i,y_i, y_i^t)\}_{i=1}^B$, where $x_i$ represents the $i$-th input image in the current mini-batch, $y_i$ represents the class label across all tasks, and $y_i^t$ represents the class label within its corresponding task $t$.
We denote the feature extracted from the foundation model as $f^b_i = F (x_i; \theta_{F_p})$, and the features extracted from the lightweight decompostion model as $f^d_i = F(x_i; \theta_{F_s};\theta_{E_t})$.
Then, the $\mathcal{L}_{\mathrm{transfer}}$ for sample $x_i$ can be written as $\mathcal{L}_{\mathrm{KL}}(f_{i}^b,f_{i}^d)$, where $\mathcal{L}_{\mathrm{KL}}$ represents the Kullback-Leibler divergence.

Moreover, we can also leverage class label information $\{y_i^t\}$ to enhance  task-level supervision. Specifically, during training, we integrate $T$ classification heads $\{h_1,...,h_T\}$ into the lightweight decompostion model (classifier in Figure~\ref{fig:method}). These classification heads can individually predict $\{Y_1,...,Y_T\}$ classes where $Y_t$ represents the  number of classes for the $t$-th task, $Y$ is the total number of all classes and $\sum_{i=1}^TY_i=Y$.
The logits extracted from the lightweight decompostion model can be denoted as $g^d_i = h_t (f^d_i)$ and the prediction probability can be calculated by
\begin{equation}
\setlength{\abovedisplayskip}{3pt}
\setlength{\belowdisplayskip}{3pt}
p_{i,j}^d = \frac{\exp(g^d_{ij}/\alpha)}{\Sigma_{j=1}^{Y_t}\exp(g^d_{ij}/\alpha)},\nonumber
\end{equation}
where $g^d_{ij}$ represents the $j$-th logit in $g^d_i$ and $\alpha$ is the temperature. $\mathcal{L}_{\mathrm{CE}}(y_i^t,p^d_{i})$ represents the task-level supervision loss of $x_i$. Then, the total loss can be written as:
\begin{equation}
\setlength{\abovedisplayskip}{3pt}
\setlength{\belowdisplayskip}{3pt}
\mathcal{L}_{\mathrm{total}}=\frac{1}{B}\sum_{t=1}^{T}\sum_{i=1}^{B^t}\left [ \mathcal{L}_{\mathrm{CE}}(y_i^t,p^d_{i})+\beta\alpha^2\mathcal{L}_{\mathrm{KL}}(f_{i}^b,f_{i}^d)\right ],\nonumber
\end{equation}
where $\beta$ is a hyperparameter.

\subsection{Task Knowledge Switch}
After decomposition, the lightweight decomposition model enables easy switching between different task knowledge through task knowledge switch, allowing for conversion to the corresponding expert model based on the requirements of various medical departments. Specifically, when deploying the model on the $t$-th task, the original parameters $\mathbf{W_0}$ can be replaced with $\mathbf{W_t} = \mathbf{W_0}+\mathbf{B_t}\mathbf{A_t}$. Similarly, when switching knowledge to another task $t^{\prime}$,  expert knowledge can be conveniently switched using $\mathbf{W_0} = \mathbf{W_t}-\mathbf{B_t}\mathbf{A_t}$ and $\mathbf{W_{t^{\prime}}} = \mathbf{W_0} + \mathbf{B_{t^{\prime}}}\mathbf{A_{t^{\prime}}} $. The parameter fusion mechanism of low-rank expert modules ensures that the deployed expert models consistently maintain a size equal to $F_s$.

\section{Experiments}

\subsection{Experimental Setup}
\par{\noindent \bf Dataset.}
To evaluate the decomposition performance, we use three medical multi-task datasets with different data scales: Radimagenet~\cite{mei2022radimagenet} (1.35 million images),  MedMnist~\cite{yang2023medmnist} (705,689 images) and  Med-MT (119,655 images). We decompose the foundation models pre-trained on these datasets into 11/10/8 lightweight expert models, respectively, with each decomposed expert model focusing on a specific anatomical region. Detailed information about these datasets can be found in the supplementary materials.

In addation, to determine the extent to which the decomposed expert models can fully replace pre-trained models in specific domains, we evaluated the transferability of these expert models on seven different downstream datasets, including COVID~\cite{xingyi2020covid_CT}, BTC~\cite{saleh2020BTC}, AD~\cite{AD}, Mura~\cite{rajpurkar2017mura}, AUITD~\cite{AUITD}, HAM10000~\cite{tschandl2018ham10000}, and DET10~\cite{liu2020chestxdet10}. Detailed information  can be found in the supplementary materials.

\par{\noindent \bf Competitive methods.}
(1) \textbf{Baseline} refers to training from scratch on downstream tasks.
(2) \textbf{Single-Task Learning (STL)} refers to training multiple single-task networks independently.
(3) \textbf{Multi-Task Learning (MTL)} refers to training a single model to predict all tasks.
(4) \textbf{STL-KD} and (5) \textbf{MTL-KD} correspond to the KD version of STL and MTL, respectively, which utilize knowledge distillation to transfer knowledge from the pre-trained models.
(6) \textbf{MoCo-MTL}~\cite{fernando2022mitigating} and (7) \textbf{Aligned-MTL}~\cite{senushkin2023independent} are the advanced MTL algorithms.
(8) \textbf{KF}~\cite{yang2022factorizing} represents the advanced knowledge decomposition method, which is the closest to our goal and serves as our primary comparison object.
{We explain the purpose of using these methods for comparison in the supplementary materials.}

\par{\noindent \bf Implementation details.} 
For the decomposition training, we use the SGD optimizer with a learning rate of 0.05 and  CosineAnnealingLR as the scheduler for training 100 epochs. For the downstream fine-tuning, we use AdamW optimizer with a learning rate of 5e-5 and train the model for 240 epochs. 
The default values for the hyperparameters are set as follows: $\alpha$=10, $\beta$=1 and $r$=8. The pre-trained model structure is ResNet50~\cite{he2016deep}, and the structure of the lightweight decomposition model is ShuffleNetV2~\cite{ma2018shufflenet}. In our experiments, we apply EKS convolution to all convolutions in ShuffleNetV2 except for GConv.

\subsection{Decomposition performance}

\begin{table*}[ht!]\footnotesize
\caption{The decomposition performance on pre-training 
 datasets. Each column represents the performance of different methods for specific tasks. It is worth noting that except for KF and ours, the concept of knowledge  decomposition does not exist in other methods. The presence of homonymous experts implies different modalities. For more details, please refer to the supplementary materials.}
% \vspace{-0.25cm}
\centering
% \resizebox{0.98\textwidth}{!}{
\resizebox{\textwidth}{!}{
\setlength{\tabcolsep}{2mm}{
\begin{tabular}{c|c|ccccccccccc|c}
% \toprule
\multicolumn{14}{c}{\textbf{Radimagenet (1.35 million images, 11 tasks)}} \\
% \midrules
\toprule[1.5pt]
Method  &{Params(M)} & Lung & Abdomen & Thyroid & Abdomen & Knee & Shoulder & Spine & Ankle & Abdomen  & Brain & Hip  & Avg \\
\midrule
{Foundation} & 23.51  & 36.22 & 46.52 & 74.05 & 48.42 & 40.09 & 31.32 & 17.79  & 12.95 & 64.17 & 77.30 & 32.33   & 43.74\\
\midrule
{STL} & 13.79 & 76.42 & 33.94 & 91.55 & \textbf{69.17} & \textbf{49.32} & 41.80 & 20.62 & \textbf{20.31} & 65.99 & 83.88 & 51.05  & 54.91\\
{MTL} & 1.25 & 77.16 & 37.45 & 91.73 & 68.43 & 46.47 & 42.72 & 20.85 & 18.17 & 71.13 & \textbf{84.67} & 55.16 & 55.81 \\
{STL-KD} & 13.79 & 78.00 & 31.74 & 91.34 & 69.10 & 46.57 & 43.09 & 19.77 & 19.43 & 69.85 & 83.83 & 52.19   & 54.99\\
{MTL-KD} & 1.25 & \textbf{78.92} & 33.89 & 91.97 & 68.54 & 48.51 & 43.34 & 21.03 & 18.48 & 69.58 & 84.18 & 54.90  & 55.75\\
MoCo-MTL & 1.25 & 76.28 & \textbf{45.56} & 86.26 & 67.00 & 45.58 & \textbf{43.97} & 18.74 & 17.41 & \textbf{74.88} & 84.33 & 52.71  & 55.70\\
Aligned-MTL & 1.25 & 77.74 & 36.38 & 91.76 & 68.51 & 48.41 & 43.28 & 21.26 & 18.37 & 68.57 & 84.54 & 54.86  & 55.79\\
\midrule
{KF} & 5.01 & 64.57 & 20.38 & \textbf{95.82} & 68.05 & 45.56 & 39.03 & \textbf{24.18} & 16.69 &56.65 & 78.46 & 51.74  & 51.01\\
{LoRKD} & 2.21 & 78.72 & 36.95 & 91.87& 68.77 & 48.80 & 43.26 & 21.41 & 19.26  & 69.24 & 84.60 & \textbf{55.93}  & \textbf{56.26}  \\
\bottomrule
\end{tabular}}}
\label{upstream}
% \vspace{-0.15cm}
\end{table*}
\begin{table*}[ht!]\footnotesize
% \caption{}
% \vspace{-0.2cm}
\centering
\resizebox{\textwidth}{!}{
% \resizebox{0.88\textwidth}{!}{
\setlength{\tabcolsep}{3mm}{
\begin{tabular}{c|c|cccccccccc|c}
% \toprule
% \midrule
\multicolumn{13}{c}{\textbf{MedMnist (705,689 images, 10 tasks)}} \\
% \midrule
\toprule[1.5pt]
Method &{Params(M)}  & Colon & Retinal & OrganC & Cell & Breast & Tissue & Skin & OrganA & OrganS & Chest   & Avg \\
\midrule
Foundation & 23.51 & 87.41 & 77.40 & 23.51 & 50.37  & 84.62 & 40.55 & 12.92 & 18.64 & 18.90 & 86.22   & 50.05\\
\midrule
STL& 12.54 & 84.53 & 78.40 & 89.65 & \textbf{96.81} & 85.26 & \textbf{68.89} & 73.97 & 92.90 & 77.43 & 85.42 &    83.33\\
MTL& 1.25 & 80.99 & 77.10 & 89.90 & 95.67 & 83.33 & 65.42 & 74.21 & 91.33 & 76.34 & 86.89 &     82.12\\
STL-KD& 12.54 & 84.33 & 77.10 & 90.45 & 96.52 & 83.33 & 68.25 & \textbf{74.81} & \textbf{93.53} & \textbf{77.52} & 82.53 &    82.84\\
MTL-KD& 1.25 & 82.83 & 75.20 & 90.02 & 95.94 & 83.26 & 64.56 & 74.31 & 92.13 & 76.02 & 86.39 &     82.06\\
MoCo-MTL& 1.25 & 76.10 & 69.80 & 80.00 & 86.55 & 76.92 & 63.89 & 69.18 & 83.82 & 67.81 & 83.87 &    75.79\\
Aligned-MTL& 1.25 & 79.78 & 73.10 & 89.70 & 95.44 & \textbf{88.46} & 64.00 & 74.36 & 90.81 & 75.06 & 86.22 &   81.69 \\
\midrule
{KF}& 4.67  & 37.83 & 48.20 & 72.40 & 44.93 & 80.13 & 54.17 & 38.01 & 71.75 & 59.19 & 72.12   & 57.87\\
{\textbf{LoRKD}}& 2.12  & \textbf{83.90} & \textbf{78.60} & \textbf{90.57} & 96.26 & 87.18 & 67.01 & 73.97  & 92.83 & 77.27 & \textbf{87.39}   & \textbf{83.50} \\
% \midrule
\bottomrule
\end{tabular}}}
% \label{upstream}
\vspace{-0.15cm}
\end{table*}
\begin{table*}[ht!]\footnotesize
% \caption{}
\vspace{-0.2cm}
\centering
% \resizebox{0.74\textwidth}{!}{
\resizebox{\textwidth}{!}{
\setlength{\tabcolsep}{4.2mm}{
\begin{tabular}{c|c|cccccccc|c}
% \toprule
% \midrule
\multicolumn{11}{c}{\textbf{Med-MT (119,655 images, 8 tasks)}} \\
% \midrule
\toprule[1.5pt]
Method &{Params(M)}  & Retinal & Skin & Breast & GI tract & Lung & Shoulder & Lung & Bone    & Avg\\
\midrule
{Foundation}& 23.51 & 81.83 & 87.01 & 81.82 & 91.25 & 66.37 & 92.31 & 65.00 & 59.46    &78.13\\
\midrule
{STL}& 10.03  & 75.27 & 77.92 & 76.59 & 85.62 & 69.91 & 75.00 & 64.85 & 51.15 &   72.04\\
{MTL}& 1.25 & 78.14 & 78.57 & 77.85 & 87.94 & 69.91 & 79.81 & 64.37 & 49.41     & 73.25\\
{STL-KD}& 10.03  & 71.45 & 67.53 & 77.18 & 86.06 & 60.18 & 78.85 & 64.67 & 51.23    & 69.64 \\
{MTL-KD}& 1.25  & 79.23 & 77.27 & 77.89 & 88.06 & 76.11 & 77.88 & 64.84 & 49.17    & 73.80 \\
MoCo-MTL& 1.25  & 58.74 & 55.84 & 51.74 & 48.31 & 67.26 & 67.31 & 46.76 & 20.11     & 52.01\\
Aligned-MTL& 1.25  & 61.07 & 56.49 & 51.50 & 52.63 & 69.03 & 67.31 & 46.77 & 19.17     & 53.00\\
\midrule
{KF} & 3.99 & 65.30 & 74.67 & 52.19 & 61.12 & \textbf{77.88} & 79.81 & 60.21 & 33.50     & 63.09\\
{\textbf{LoRKD}} & 1.95 & \textbf{79.37} & \textbf{85.06} & \textbf{79.04} & \textbf{88.63} & 72.57 & \textbf{83.65}& \textbf{65.07}  & \textbf{52.42}     & \textbf{75.73}\\
\bottomrule[1.5pt]
% \vspace{-0.2cm}
\end{tabular}}}
% \label{upstream}
\vspace{-0.2cm}
\end{table*}

The performance comparison of different methods on three pre-training datasets is shown in Table~\ref{upstream}. 
Each column corresponds to a specific task. ``Avg" represents the task-level average accuracy.  ``Parmas" represents the total number of parameters during training. Only KF and our method focus on the knowledge decomposition of pre-trained models.
Considering the generalization requirement of the foundation model, it is common for the foundation model to use a unified classification head during training, instead of configuring a specific classification head for each task~\cite{mei2022radimagenet}. This is also why the performance of the foundation model in Table~\ref{upstream} is relatively poor. Note that other methods can only solve problems within specific tasks.

\par{\noindent \bf The foundation model vs. STL.} The results of the foundation model are superior to STL on Med-MT, but significantly inferior to STL on Radimagenet and MedMnist, particularly for MedMnist, which suggest that as the increase of scale and diversity of the pre-training dataset, the specialization of the pre-trained model gradually diminishes due to conflicts between different domain knowledge. In contrast, training models independently for each task (STL) can prevent interference between different tasks and thus achieve better performance than foundation models on Radimagenet and MedMnist. However, STL is unable to learn common knowledge among tasks, so it often requires more data to ensure its generalization ability. Besides, training $T$ individual models is not only time-consuming but also results in a linear increase of parameters.

\par{\noindent \bf MTL-based methods vs. STL-based methods.} We can also observe that MTL outperforms STL in Radimagenet and Med-MT, while underperforms STL in MedMnist. It may be related to the degree of correlation between tasks included in the pre-training dataset, where MedMnist has the most diverse modalities (refer to supplementary materials). In contrast to the standard MTL, other advanced MTL methods, namely MoCo-MTL and Aligned-MTL, do not yield improvements and may even exhibit worse performance. This observation suggests that balancing multiple optimization objectives to obtain a better shared encoder is not an effective solution when there are significant differences among tasks.
The knowledge distillation variants of STL and MTL (STL-KD and MTL-KD) do not show significant performance improvement, which suggests that the general features extracted by foundation models have limited benefits for specific tasks and indirectly reflects the importance of both specialized and general features. It aligns with the design philosophy of our LoRKD.

\par{\noindent \bf LoRKD vs. KF and other methods.} Compared to the knowledge decomposition method KF, 
our method shows significant performance advantages and introduces fewer parameters. Specifically, even including 11/10/8 experts, our method has less than half the number of parameters of KF.
This result validates the effectiveness of our low-rank expert modules and the efficient knowledge separation convolution. Furthermore, our method also achieves the best average performance compared to other non-knowledge decomposition baselines, highlighting the potential of knowledge decomposition in extracting task-specific knowledge.

% \newpage
\subsection{Transferability}
\label{Transferability}

\begin{table*}[th]\footnotesize
\caption{The performance of the decomposed expert models on seven downstream datasets. ``Params" represents the number of model parameters during deployment. ``Comp. Ratio" denotes the compression ratio, defined as the ratio of the deployed model parameters to the parameters of the foundation model. ``-" indicates the absence of data corresponding to the downstream tasks in the pre-training dataset.}
% \vspace{-0.2cm}
\centering
\resizebox{0.96\textwidth}{!}{
\begin{tabular}{c|c|cc|c|c|c|c|c|c|c|c}
\toprule[1.5pt]
{Pre-train} & {Model}& {Params(M)} & {Comp. Ratio} & {COVID} & {BTC} & {AD} & {Mura\_s} & {AUITD} & {HAM10000} & {DET10}& {Avg}  \\
% \cmidrule(r){4-15} 
% &  &  & ACC& AUC& ACC& AUC&ACC &AUC &ACC &ACC  &  ACC   \\
\midrule
\multirow{11}{*}{\rotatebox[origin=c]{90}{Radimagenet}} &Foundation  & 23.51 & / & 78.33 & 80.20 & 74.35 & 71.05 & 96.66 & 75.08 & 86.69 & 80.34 \\
% \midrule
\cmidrule(r){2-12} 
&Baseline  & 1.25 & 5.32\% & 82.76 & 75.38 & 76.08& 76.73 & 97.77 & 74.42  & 87.54 & 81.52 \\
&STL & 1.25 & 5.32\% & 82.76 & 78.93 & 76.70 & 77.26 & 97.77 & - & 87.52 & - \\
&MTL & 1.25 & 5.32\% & 83.25 & 79.95 & 74.67 & 76.91 & 97.77 & 75.83 & 86.82 & 82.17 \\
&STL-KD & 1.25 & 5.32\% & 82.27 & 80.46 & 76.31 & 76.73 & 96.66 & - & 87.25 & - \\
&MTL-KD & 1.25 & 5.32\% & 81.77 & 78.93 & 73.89 & 76.55 & 96.66 & 74.37 & 87.17 & 81.33 \\
&MoCo-MTL & 1.25 & 5.32\% & 78.82 & 78.68 & 69.27 & 75.49& 91.64 & 71.77  & 86.43 & 78.87 \\
&Aligned-MTL & 1.25 & 5.32\% & 82.27 & 78.43 & 70.29 & 76.91 & 88.58 & 73.07 & 86.91 & 79.49 \\
\cmidrule(r){2-12} 
&KF & 1.60 & 6.81\% & 80.79 & 79.70 & 71.23 & 74.96 & 96.66 & $74.12^{\dagger}$ & 87.17 & 80.66 \\
&\textbf{LoRKD} & 1.25 & 5.32\% & \textbf{86.21} & \textbf{81.47} & \textbf{79.12} & \textbf{79.57} & \textbf{98.33} & $\textbf{76.03}^{\dagger}$ & \textbf{88.50} & \textbf{84.18} \\
\midrule
\midrule
\multirow{11}{*}{\rotatebox[origin=c]{90}{MedMnist}}  &Foundation  & 23.51 & / & 80.30 & 77.41 & 72.09 & 76.38 & 88.86 & 72.12 & 86.80 & 79.14 \\
\cmidrule(r){2-12} 
&Baseline  & 1.25 & 5.32\% & 82.76 & 75.38 & 76.08& 76.73 & 97.77 & 74.42  & 87.54 & 81.52 \\
&STL & 1.25 & 5.32\% & 83.25 & - & - & - & 97.77 & 71.82 & 87.56 & - \\
&MTL & 1.25 & 5.32\% & 81.28 & 78.68 & 77.17 & 76.19 & 97.77 & \textbf{74.82} & 87.36 & 81.90 \\
&STL-KD & 1.25 & 5.32\% & 79.80 & - & - & - & 97.49 & 73.87 & 86.93 & - \\
&MTL-KD & 1.25 & 5.32\% & 80.79 & 78.62 & 76.62& 75.84 & 98.05 & 73.87 & 87.23 & 81.57  \\
&MoCo-MTL & 1.25 & 5.32\% & 78.82 & 77.16 & 77.80 & 74.95 & 97.77 & 72.77 & 86.82 & 80.87 \\
&Aligned-MTL & 1.25 & 5.32\% & 82.27 & 77.42 & 77.72 & 76.90 & 96.37 & 73.87 & 86.97 & 81.65 \\
\cmidrule(r){2-12} 
&KF & 1.60 & 6.81\% & 80.79 & $77.15^{\dagger}$ & $72.71^{\dagger}$ & $74.77^{\dagger}$ & 96.10 & 72.97 & 87.41 & 80.27 \\
&\textbf{LoRKD} & 1.25 & 5.32\% & \textbf{84.24} & $\textbf{79.70}^{\dagger}$ & $\textbf{79.05}^{\dagger}$ & $\textbf{77.80}^{\dagger}$ & \textbf{98.33} & \textbf{74.82} & \textbf{87.60} & \textbf{83.08} \\
\midrule
\midrule
\multirow{11}{*}{\rotatebox[origin=c]{90}{Med-MT}}  &Foundation  & 23.51 & / & 82.76 & 78.17 & 69.19 & 71.76 & 89.69 & 75.53  & 86.69 & 79.11 \\
\cmidrule(r){2-12} 
&Baseline  & 1.25 & 5.32\% & 82.76 & 75.38 & 76.08& 76.73 & 97.77 & 74.42  & 87.54 & 81.52 \\
&STL & 1.25 & 5.32\% & - & - & - & - & - & 73.77 & - & - \\
&MTL & 1.25 & 5.32\% & 82.76 & 76.65 & \textbf{77.48} &77.09 & 97.49 & 74.92 & 87.15 & 81.93 \\
&STL-KD & 1.25 & 5.32\% & - & - & - & - & - & 74.42 & - & - \\
&MTL-KD & 1.25 & 5.32\% & 82.76 & 75.89 & 74.43 & 76.91 & 97.77 & 74.32 & 87.34 & 81.34 \\
&MoCo-MTL & 1.25 & 5.32\% & 80.79 & 76.40 & \textbf{77.48} & 76.91 & 97.49 & 72.62 & 86.91 & 81.23 \\
&Aligned-MTL & 1.25 & 5.32\% & 79.80 & 75.63 & 76.62 & 76.73 & 97.77 & 73.72 & 87.19 & 81.06 \\
\cmidrule(r){2-12} 
&KF & 1.60 & 6.81\% & $80.79^{\dagger}$ & $74.87^{\dagger}$ & $75.76^{\dagger}$ & $76.73^{\dagger}$ & $98.05^{\dagger}$ & 73.92 & $87.39^{\dagger}$ & 81.07 \\
&\textbf{LoRKD} & 1.25 & 5.32\% & $\textbf{83.25}^{\dagger}$ & $\textbf{77.66}^{\dagger}$ & $76.94^{\dagger}$ & $\textbf{78.33}^{\dagger}$ & $\textbf{98.33}^{\dagger}$ & \textbf{75.18} & $\textbf{87.84}^{\dagger}$ & \textbf{82.50} \\
\bottomrule[1.5pt]
\end{tabular}}
\label{downstream}
\vspace{-0.3cm}
\end{table*}

\begin{table}[t]
    \centering
    \caption{The impact of $r$ on performance and transferability.}
    \vspace{-0.2cm}
    \resizebox{0.48\textwidth}{!}{
    \begin{tabular}{c|c|ccccc|c}
    \toprule[1.5pt]
        & Pre-train & COVID & BTC & AD & Mura\_s & AUITD  & Avg\\
    \midrule
      $r$=4 & 55.08  & 85.71  & 79.95 &  75.61 & 77.98  &98.05  &83.46\\
      % \hline
      $r$=8 & 56.26  & 85.71  & 81.47 &  75.92 & 78.51  &98.33  &84.93\\
      % \hline
      $r$=16 & 56.19  & 86.21  & 82.49 &  78.81 & 78.51  &98.33  &84.87\\
    \bottomrule[1.5pt]
    \end{tabular}}
    \label{tab:rank_proformance}
    \vspace{-0.4cm}
\end{table}

In order for the decomposed lightweight expert model to fully replace the foundation model within a specific domain, it is necessary not only for the expert models to perform well on the same distribution of data (pre-training dataset), but also to evaluate its generalization ability on downstream tasks with close distributions.
The performance comparison of the expert models decomposed from three pre-training datasets on seven downstream datasets is shown in Table~\ref{downstream}. 

For KF and our method, we fine-tune the corresponding expert models on downstream datasets, such as using the lung expert model for the COVID dataset. 
If there was no corresponding expert model, similar to~\cite{yang2022factorizing}, we fine-tune on the shared backbone (with $^{\dagger}$). 
As for other non-knowledge decomposition methods, we use the models trained on the pre-training dataset for fine-tuning to demonstrate the advantages of knowledge decomposition in terms of transferability. For details, please refer to supplementary materials.

The performance of fine-tuning foundation models is observed to be inferior to Baseline, providing evidence again that the foundation model cannot replace task-specific models in terms of performance, due to lack of specialization.
Compared to baseline, both STL-based and MTL-based methods show little improvement, indicating that solely focusing  on task-specific knowledge or  common knowledge does not contribute to transferability. 
Conversely, our expert models incorporate both common knowledge and task-specific knowledge, which exhibit strong transferability and even significantly outperform KF.
Another benefit when compared to KF is that our method is compatible with the parameter fusion and does not require the simultaneous deployment of two networks (CKN and the corresponding TSN need to be deployed simultaneously in KF).

Furthermore, we discover an interesting phenomenon. In comparison to MTL-KD, our method outperforms it more significantly on downstream datasets. 
This shows the advantage of knowledge decomposition in transferability, and the transferability can not be directly reflected through the decomposition performance.
And as the scale of the pre-training dataset increases, the transferability of our decomposed expert models also improves, indicating that increasing the scale of pre-training datasets benefits the transferability of the decomposed model.

\subsection{Ablation Studies and More Analysis}

\par{\noindent \bf The impact of Rank $r$.}
We evaluate the impact of $r$ for decomposition performance on pre-training dataset and transferability on downstream datasets in Table~\ref{tab:rank_proformance}.
The results show that for Radimagenet, increasing $r$ from 4 to 8 leads to a significant performance improvement, while increasing $r$ from 8 to 16 does not provide further enhancement. This suggests that selecting a larger $r$ is not necessarily better and may be related to the scale of datasets.
Moreover, we find that the improvement on pre-training dataset is positively correlated with the improvement in transferability.

\begin{table}[t]
    \centering
    \caption{The comparison of the costs of different methods.}
    \vspace{-0.2cm}
    \resizebox{0.48\textwidth}{!}{
    \begin{tabular}{c|c|c|ccc|c}
    \toprule[1.5pt]
        \multirow{2}{*}{Costs} & \multirow{2}{*}{Baseline}  & \multirow{2}{*}{KF}  & \multicolumn{3}{c|}{\textbf{LoRKD}}   & \multirow{2}{*}{Foundation}\\
        \cmidrule(r){4-6} 
        & & & r=4 &  r=8 & r=16 &\\
    \midrule
      % \multicolumn{7}{c}{During training} \\
      % \midrule[1.2pt]
      Params (Training)& 1.25  & 5.01  & 1.73 &  2.21 & 3.16  &23.51\\
      FLOPs (Training)&  0.15 &  0.63& 0.21 &  0.27 & 0.38  &4.13\\
      \midrule
      % \multicolumn{7}{c}{After deployment} \\
      % \
        Params (Deployment)& 1.25  & 1.60  & 1.25 &  1.25 & 1.25  &23.51\\
      FLOPs (Deployment)&  0.15 &  0.16& 0.15 &  0.15 & 0.15  &4.13\\

    \bottomrule[1.5pt]
    \end{tabular}}
    \label{tab:cost}
    \vspace{-0.5cm}
\end{table}

\begin{figure*}[t]
    \centering
    \includegraphics[width=0.98\linewidth]{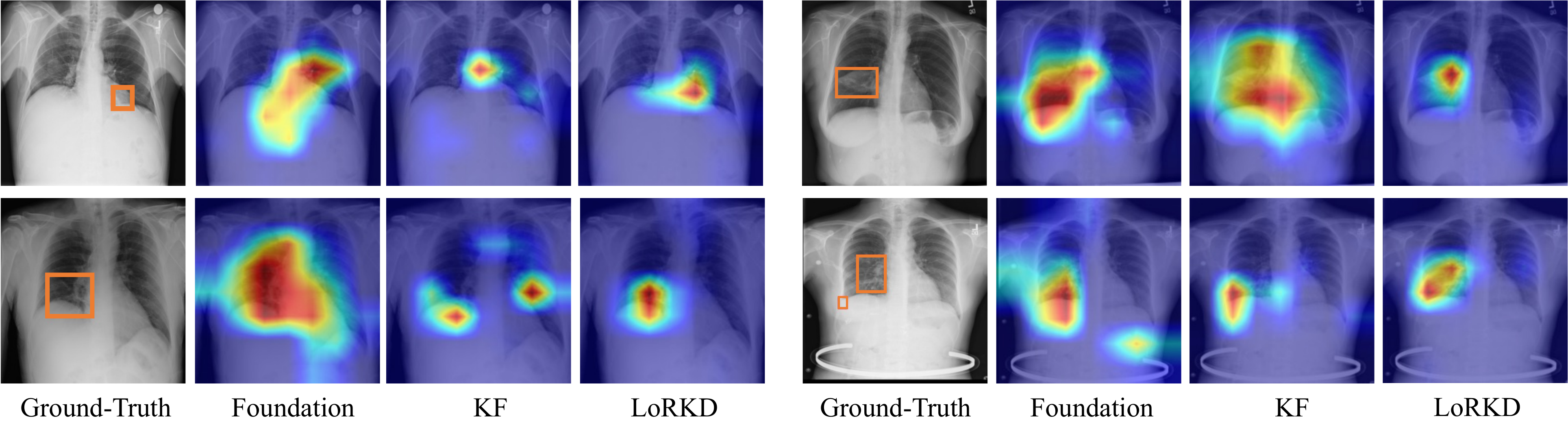}
    \vspace{-0.1cm}
    \caption{Comparison of Grad-CAM visualizations between the decomposed model and the foundation model on DET10. The foundation model tends to focus on larger regions, corresponding to its general feature extraction capability, while our decomposed expert model focuses on more precise regions, reflecting stronger specialization.}
    \label{fig:grad_cam}
    \vspace{-0.23cm}
\end{figure*}

\par{\noindent \bf Lower Costs {and Higher Efficiency.}}
We compare the number of parameters and the FLOPs  among different methods on Radimagenet, as shown in Table \ref{tab:cost}.
It can be observed that, compared to the foundation model, ours significantly reduce the number of parameters and FLOPs, demonstrating that our method can effectively reduce  deployment costs while maintaining high computational efficiency. 
In comparison to KF, even at $r$=16, our method still incurs significantly lower costs. 
As $r$ increases, our costs do not increase significantly and remains at a lower level.
It is worth mentioning that if parameter fusion is used, our costs will be the same as baseline. 
% \par{\noindent \bf More Efficiency.}
% ~\cite{hu2021lora}
{
Furthermore, we compare the efficiency of different methods in Table~\ref{tab:r1q1} following the setting in~\cite{wen2023batched}, where $b$, $r$, $l$ and $d$ represent the batch size, the rank, the maximum sequence length and the feature dimension respectively. $c_2$ is the computational coefficient of matrix multiplication.
If the following condition is satisfied,
 $\frac{rbld^2}{Td^2r+bd^2l} \ge 1  \Longrightarrow  \frac{Tr}{bl} + 1\le r$, the efficiency of EKS conv is higher. This inequality holds true in most real-world cases, as $bl > Tr$, $r > 2$ are common settings.  A detailed explanation is provided in the supplementary materials.}
% the supplementary materials such as~\cite{wen2023batched}, theoretically indicating that our method is more efficient under real-world conditions.

\begin{table}[t]
    \centering
    % \vspace{-0.15cm}
    \caption{Efficiency comparison of different methods for constructing personalized low-rank experts for each sample.}
    \resizebox{0.48\textwidth}{!}{
    \setlength{\tabcolsep}{1mm}{
    \begin{tabular}{c|c|c}
    \toprule[1.5pt]
        Method & Improved Operation & Computational Cost\\
        % \midrule
        %   LoRA & $\mathbf{Y}={\mathbf{X}W_0}+\varphi(\varphi(\mathbf{X},\mathbf{B}),\mathbf{A})$ &$2c_1(dblr) + c_2(bld^2)$\\
         \midrule
         FLoRA~\cite{wen2023batched} & $\mathbf{Y}=\mathbf{A}\circ\left((\mathbf{B}\circ\mathbf{X})\mathbf{W_0}\right)$&$c_2(rbld^2)$\\
         \midrule
         EKS conv (ours) & $\mathbf{Y}=\mathbf{X}(\mathbf{W_0}+\sum\nolimits_{i=1}^{T} (\mathbf{\widetilde{BA}}\odot \mathbf{M})_i)$&$Tc_2(rd^2)+c_2(bld^2)$\\
    \bottomrule[1.5pt]
    \end{tabular}}}
    % \caption{Caption}
    \label{tab:r1q1}
    \vspace{-0.35cm}
\end{table}

\par{\noindent \bf Stronger Specialization.}
Taking the DET10 dataset with detection annotations as an example, we evaluate the differences in the activated regions between the decomposed expert model and the foundation model during the prediction process from the perspective of Grad-CAM~\cite{selvaraju2017grad}. The corresponding visualization results are shown in Figure \ref{fig:grad_cam}.

It can be observed that although the foundation model can focus on the correct regions, the range of regions it attends to is usually larger compared to the detection boxes of Ground-Truth. 
This may be due to the fact that the feature extractor of the foundation model retains a certain degree of general feature extraction ability, tending to focus on more information regardless of the specific task. 
Conversely, our expert models focus on more precise abnormal regions and demonstrate stronger specialization compared to the foundation model. 
Compared to KF, our approach also achieves higher recognition accuracy, further indicating that our decomposed model has better transferability.

\par
\begin{figure}[t]
    \centering
    \includegraphics[width=0.98\linewidth]{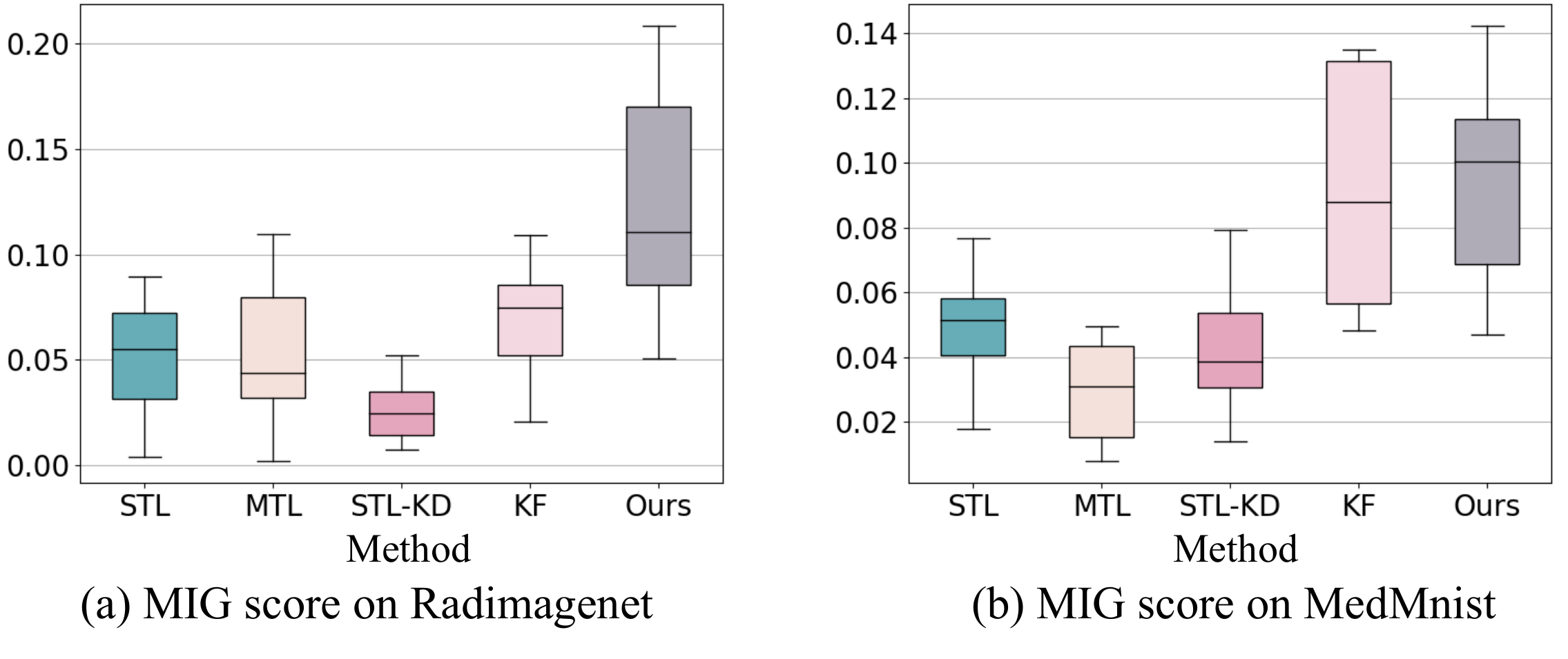}
    \vspace{-0.22cm}
    \caption{The comparison of MIG scores on different methods.}
    \label{fig:mig}
    % \vspace{-0.5cm}
    \vspace{-0.35cm}
\end{figure}

{\noindent \bf Knowledge Disentanglement.} 
To verify whether our method can indeed achieve knowledge decoupling between different tasks, we measure the mutual information gap (MIG) scores~\cite{chen2018isolating} for different methods. MIG is a widely used metric for quantifying disentanglement. The results are shown in Figure \ref{fig:mig}, where higher MIG scores indicate a higher level of disentanglement. It can be observed that our method exhibits a higher level of disentanglement compared to the previous KF and other baselines, which may benefit from our explicit gradient separation.

In addition, we find that the degree of disentanglement in MTL is lower compared to STL. This suggests that when MTL employs a shared encoder to acquire common knowledge for across tasks, the entanglement of gradients from different tasks 
also results in knowledge entanglement. 
Additionally, compared to STL, the degree of disentanglement in STL-KD is also lower, which can be attributed to the transfer of common knowledge from the foundation model.

{More analysis, such as using larger foundation models and comparing CKA feature similarity across different tasks, can be found in the supplementary materials.}

\section{Conclusion}
This paper proposes a new perspective called knowledge decomposition, which focuses on reducing the deployment costs and enhancing specialization for medical foundation models. We design a low-rank expert module and an efficient gradient separation convolution to successfully decompose the foundation model into multiple lightweight expert models, and validate their transferability and  disentanglement. We hope that this research will provide new insights for the development of medical foundation models.

\noindent\textbf{Acknowledgment.} This work is supported by the National Key R\&D Program of China (No. 2022ZD0160703),  STCSM (No. 22511106101, No. 22511105700, No. 21DZ1100100), 111 plan (No. BP0719010) and National Natural Science Foundation of China (No. 62306178).
\clearpage
{
    \small
    \bibliographystyle{ieeenat_fullname}
    \bibliography{main}
}


\end{document}